\documentclass[11pt]{article}
\usepackage[preprint]{acl}
\usepackage{microtype}

\usepackage[english,bidi=default]{babel}
\babelfont{rm}{TeXGyreTermesX}
\usepackage{fontspec}
\newfontfamily\arabicfont[Path=./,Script=Arabic]{NotoSansArabic-Regular.ttf}
\newfontfamily\devanagarifont[Path=./,Script=Devanagari]{NotoSansDevanagari-Regular.ttf}
\newfontfamily\tamilfont[Path=./,Script=Tamil]{NotoSansTamil-Regular.ttf}
\newfontfamily\cjkfont[Path=./]{NotoSansCJKsc-Regular.otf}
\newfontfamily\cyrillfont[Path=./]{NotoSerif-Regular.ttf}
\newfontfamily\greekfont[Path=./]{NotoSerif-Regular.ttf}
\newcommand{\ar}[1]{{\arabicfont #1}}
\newcommand{\hi}[1]{{\devanagarifont #1}}

\newcommand{\zh}[1]{{\cjkfont #1}}
\newcommand{\cyr}[1]{{\cyrillfont #1}}
\newcommand{\grk}[1]{{\greekfont #1}}

\usepackage{booktabs}
\usepackage{graphicx}
\usepackage{amsmath}
\usepackage{amssymb}
\usepackage{multirow}
\usepackage{xcolor}
\usepackage{url}
\usepackage{enumitem}
\usepackage{makecell}
\usepackage{placeins}
\usepackage{float}
\setlist{nosep,leftmargin=*}

\newcommand{\lpr}{\textsc{lpr}}

\newcommand{\indicator}{\mathbf{1}}
\newcommand{\addref}[1]{\iffalse ADDREF:#1 \fi}

\title{AI-Associated Lexical Shifts Across 34 Languages:\\Cross-Lingual Convergence and Diachronic Uptake in News Writing}

\author{
Thomas Stephan Juzek\thanks{\hspace{-0.0cm}Code:\ \href{https://github.com/tjuzek/ai-34-languages}{github.com/tjuzek/ai-34-languages}; AI Word Explorer:\ \href{https://www.aiwordexplorer.com/}{aiwordexplorer.com/}.} \\
Florida State University \\
\texttt{tjuzek@fsu.edu} \\
}

\begin{document}
\maketitle

\begin{abstract}
AI-associated lexical shifts have been documented mainly in Scientific English. We extend this work to 34 languages in the WMT News Crawl corpus, refining a split-halves continuation diagnostic that compares GPT-4.1 continuations with matched human gold-standard text. For each language, we derive ranked AI-overused lemmas using log prevalence ratios. We find substantial cross-lingual semantic convergence:\ semantically related concepts recur across typologically diverse languages, with \emph{emphasize}-type verbs appearing in 24 of 34 languages. Embedding-based and manual analyses support this pattern. We also examine diachronic uptake in news writing before and after ChatGPT's release. Tracking each language's top 20 AI-overused items, we find prevalence increases in 26 of 34 languages from 2020--2021 to 2023--2024, with a mean change of +15.1\,\%, whilst matched baseline words show no comparable increase ($-4.5\,\%$). In 10 languages with longer historical coverage, longitudinal analyses show post-2022 increases that exceed the modest shifts observed in earlier periods, though with smaller effect sizes than in Scientific English. We validate our approach extensively, including across seeds, model variants, data sizes, model families, and more. Our findings are consistent with the view that AI-associated lexical preferences extend beyond English and may exert cross-lingual homogenising pressure on global language use. 
\end{abstract}

\section{Introduction}
\label{sec:intro}
Transformer-based Large Language Models (LLMs), trained primarily via next-word prediction and later adapted for assistant use (`chat'), have seen fast uptake for writing and other tasks \cite{vaswani2017attention,wolf2020transformers,ilia2024predict}. Against this backdrop, rapid large-scale shifts in language use have been observed and are so far best documented in Scientific English, where words such as \emph{delve}, \emph{underscore}, and \emph{intricate} show sudden anomalous spikes \cite{geng2024chatgpt,liang2024mapping,liang2024monitoring,matsui2024delving,bao2025examining,juzek2025does,kobak2025delving}. Population-level analyses suggest that substantial fractions of recent scientific output are now LLM-modified \cite{coffey2024-ihe-oup,liang2025quantifying,he2026academic,thelwall2026have}. Related shifts have also been observed in English news media \cite{hanley2024machine}, spoken communication \cite{yakura2024empirical,anderson2025model}, political speech \cite{mofaddel2026generative}, and organisational writing \cite{liang2025widespread}. LLMs are also increasingly used for downstream analysis tasks such as public-opinion classification \cite{liebeskind2024opinionidentification,babadfalk2025chatgptcapabilities}. 

However, this literature remains largely English-focused and domain-specific. Existing multilingual work has mainly centred on detection of human- vs machine-generated text \cite{lavergne2008detecting,liang2023gpt,macko2023multitude,mitchell2023detectgpt,sadasivan2023can,weber2023testing,zaitsu2023distinguishing,li2024enhancedfakenews,li2024mage,schaaff2024classification,wang2024m4}; systematic characterisation and analysis of AI-associated patterns in human language use across languages remain limited.  

We address this gap by analysing 34 languages in the WMT News Crawl corpus. We refine a split-halves continuation design from the literature \citep{juzek-etal-2026-fully}, and use log prevalence ratios to derive ranked inventories of AI-overused lemmas from GPT-4.1 continuations relative to matched human text. Our findings are fourfold. First, chat-aligned models exhibit a cross-linguistically coherent lexical fingerprint:\ similar semantic concepts emerge as overused across typologically diverse languages. Second, these recurrent similarities are consistent with cross-linguistic homogenisation pressure. Third, these AI-associated words show significant diachronic uptake in real-world news text, whereas matched baseline words do not. The magnitude of change is remarkable, but smaller than the disruptions reported for Scientific English. Fourth, the signal is robust across a range of conditions, including random seeds, model sizes, model versions, data volumes, and model families. Diagnosing such divergences is a first step toward mitigation, and we release the pipeline as a reusable diagnostic. 

The paper is structured as follows. \S\ref{sec:background} reviews related work. \S\ref{sec:methods} introduces the split-halves continuation design, \S\ref{sec:metrics} defines the \lpr{}-based ranking procedure, \S\ref{sec:results} reports the main findings, and \S\ref{sec:discussion} and \S\ref{sec:conclusion} discuss the implications and conclude the paper.

\section{Background and Related Work}
\label{sec:background}

\paragraph{AI-associated lexical shifts in scientific writing.}
Scientific English is undergoing a notable lexical shift, in which words such as \emph{delve}, \emph{underscore}, and \emph{intricate} have spiked since 2022 \cite{liang2024mapping,liang2024monitoring,liu2024towards,matsui2024delving,masukume2024impact,picazo2024analysing,juzek2025does,kobak2025delving,kousha2025much,geng2026beyond,botes2025initial}. This overuse departs from historical baselines \cite{kobak2025delving,liang2025quantifying}, and many of the same words are overused in AI-generated text \cite{juzek2025does}. The pattern also appears to evolve:\ whilst high-visibility markers may decline after public scrutiny \cite{leiter2024nllg}, other LLM-favoured terms continue to rise \cite{geng2025detectability}. Furthermore, the analysis of AI-associated shifts connects to the broader literature on diachronic lexical semantic change, which examines how word meanings evolve over time \cite{hamilton2016diachronic,tahmasebi2018survey,schlechtweg2020semeval}. Existing analyses of AI-associated shifts (which concern prevalence rather than meaning) have, however, often relied on manually selected or filtered lists, which makes it harder to extend them across languages.

\begin{figure}[t]
    \centering
    \includegraphics[width=\columnwidth]{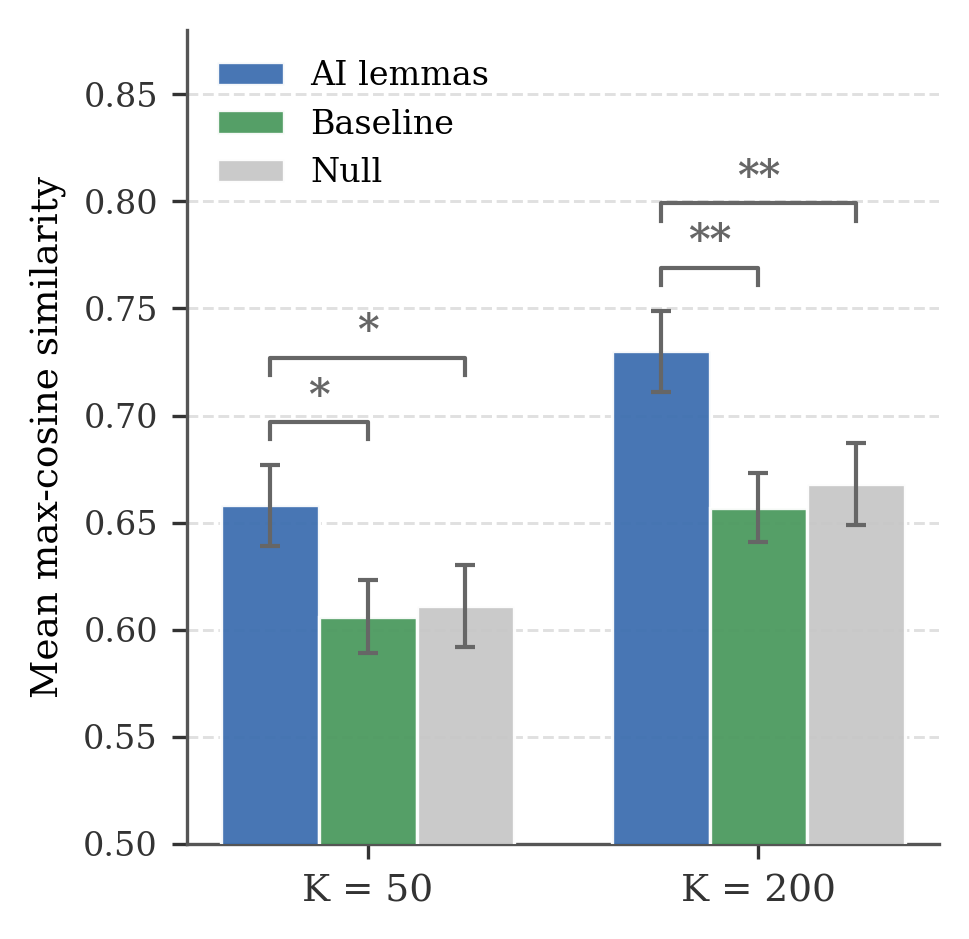}
    \caption{Embedding-based analysis of cross-lingual semantic convergence. AI-overused lemmas are more semantically similar across languages than matched baseline and null sets for top-50 and top-200 lemma lists. Error bars show 95\,\% confidence intervals; asterisks mark pairwise significance.}
    \label{fig:cosinehook}
\end{figure}

\paragraph{Beyond scientific writing.}
Evidence suggests that lexical shifts are not confined to academic writing. Since ChatGPT's release, shifts towards AI-preferred words have also been observed in spoken communication, including unscripted spoken English \cite{yakura2024empirical,anderson2025model}. More broadly, AI tools may accelerate latent trends towards specific stylistic norms \cite{rudnicka2023can,rudnicka2025each}, and algorithmic suggestions can steer writing towards more predictable forms \cite{arnold2020predictive,hohenstein2023artificial}. This raises the possibility of homogenisation pressures on content diversity \cite{moon2025homogenizing,padmakumar2023does,doshi2024generative,zhang2025verbalized}, and of effects on creativity \cite{anderson2024homogenization,kumar2025human,wenger2025we} and on cross-cultural expression \cite{agarwal2025ai,sourati2025homogenizing,utami2026can}. Incorporation of AI-influenced text into future training data may further amplify these effects \cite{shumailov2023curse,shumailov2024ai,zhang2025verbalized}. LLM-generated content has also been detected at scale in English news \cite{hanley2024machine}, Italian news \cite{puccetti2024ai}, and broader English web ecosystems \cite{sun2025we}, as well as reported in professional workplace writing \cite{liebscher2026workslop}. Likewise, a study of English news found an increase in LLM-style vocabulary, but evidence of homogenisation remains unclear \cite{fitterer2025testing}.

\paragraph{Mechanisms.}
State-of-the-art chat assistants typically undergo several training stages:\ pretraining, instruction tuning, preference learning, and task-specific fine-tuning. Whereas pretraining provides broad statistical knowledge of language and instruction tuning improves assistant-like responsiveness, preference-learning methods such as Reinforcement Learning from Human Feedback and related optimisation frameworks further steer outputs towards responses preferred by human raters \cite{christiano2017deep,ziegler2019fine,stiennon2020learning,ouyang2022training,bai2022training,bai2022constitutional,rafailov2023direct}. However, this stage optimises not only for correctness or usefulness, but also for properties that human annotators tend to reward.

Recent work shows that preference-based alignment can introduce systematic biases, including preference collapse \cite{xiao2024algorithmic}, verbosity bias \cite{saito2023verbosity,wu2025style}, sycophancy \cite{sharma2023towards,wei2023simple}, and reduced output diversity \cite{kirk2023understanding,murthy2025one,zhang2025lists,zhang2025verbalized}. More broadly, alignment may reward stylistic properties that are socially legible as helpful, safe, or sophisticated, and thereby reinforce a distinctive assistant register characterised by hedging, positivity, and recurrent lexical choices \cite{gabriel2020artificial,durmus2023towards,santurkar2023whose,he2024whose,norhashim2024measuring,young2024role,bharadwaj2025flattery,juzek2025does,chooi2026stylistic,huang2026pen}. 

\paragraph{The multilingual gap.}
Despite the global adoption of LLMs, research on their linguistic impact \cite{bommasani2021opportunities,coffey2024-ihe-oup,stackoverflow2024-ai,apnews2025-apnorc-usage,pew2025-chatgpt-usage} remains predominantly English-centric. Existing non-English and multilingual work has focused primarily on detection, on classifying individual text instances as human- or machine-generated \cite{gehrmann2019gltr,chakraborty2023possibilities,jakesch2023human,kirchenbauer2023reliability,kirchenbauer2023watermark,liang2023gpt,macko2023multitude,mitchell2023detectgpt,sadasivan2023can,weber2023testing,wang2024m4,irrgang2024features,kotz2024analisis,li2024mage,schaaff2024classification,silva2024densidade,zaitsu2023distinguishing,huang2025magret,jin2025llm,jin2025trapdoc}. Yet detection alone does not show whether or how human language changes under AI influence, nor does it identify AI-associated lexical overuse systematically across languages. A systematic cross-linguistic account of AI-mediated lexical shift remains lacking.

\section{Data and Methods}
\label{sec:methods}

\subsection{Corpus and Languages}
\label{sec:corpus}

Our data source is the WMT News Crawl corpus \cite{wmt_newscrawl}, a large-scale collection of monolingual news text in many languages. The languages are independent of each other (i.e.\ not parallel or translated). The datasets consist of year-partitioned files of shuffled, deduplicated sentences; sentences are therefore the largest coherent unit available.

We contrast pre-ChatGPT-release data (2020--2021) with post-ChatGPT-release data (2023--2024). Languages were selected by two criteria:\ sufficient WMT coverage for all four target years and availability of Stanza \cite{qi2020stanza} models for tokenisation, lemmatisation, and Universal POS (UPOS) tagging \cite{zeman2018conll,nivre2020universal}. This gives 34 languages:\ Afrikaans, Arabic, Bulgarian, Chinese, Croatian, Czech, Dutch, English, Estonian, Finnish, French, German, Greek, Hindi, Icelandic, Indonesian, Italian, Japanese, Kazakh, Korean, Kyrgyz, Latvian, Lithuanian, Marathi, Persian, Polish, Portuguese, Romanian, Russian, Serbian, Spanish, Tamil, Turkish, and Ukrainian.

Hungarian was excluded because markup and code contamination produced artefactual growth patterns in preliminary analyses. The remaining languages passed manual quality checks.

\subsection{Preprocessing}
\label{sec:preprocessing}

For the four year target period, WMT News Crawl contains about 1B sentences across the 34 languages. For computational feasibility (esp.\ regarding the POS-tagging), we cap sampling at 3M lines per language-year using a fixed seed, and retain 298M lines, with 7.1B tokens, in total. Lines exceeding 200 word tokens or 1500 characters are excluded, as they are unlikely to be single sentences. All datasets are processed with Stanza for tokenisation, lemmatisation, and Universal POS (UPOS) tagging. Our primary unit of analysis is the \emph{lemma+UPOS} key (e.g.,\ \texttt{delve\_VERB}), which collapses inflectional variants and partially disambiguates homographs (e.g.,\ \texttt{patient\_NOUN} vs \texttt{patient\_ADJ}). We include \textsc{sym} in the calculations but filter it out in reporting. By contrast, \textsc{punct} is excluded from prevalence calculations, because human continuations typically contain exactly one terminal punctuation mark, whereas model continuations may show greater variation.

\subsection{Split-Halves Comparison}
\label{sec:splithalves}

To identify AI-associated lexical items, we use a split-halves methodology that compares model continuations with matched human gold-standard continuations in the same context \citep{juzek-etal-2026-fully}. This derives scalable AI-associated lexical preferences from matched model--human behaviour. The setup is related to cloze-style evaluation paradigms \cite{ippolito2019comparison,eisape2020cloze,giulianelli2023comes}, though these typically focus on single words.

\paragraph{Prompt construction.}
For each sentence, we split the text into a first half (\emph{prompt}) and a second half (\emph{human gold continuation}). Sentences are eligible only if both halves contain at least 12 parsed tokens each. Split points are chosen near the sentence midpoint. To map token-level splits back to character-space text, we use a staged procedure:\ strict token-to-text alignment where possible, local seam search within a small window around the split when alignment fails, and reconstruction from token forms as a fallback. The reconstruction step includes language-specific handling for clitics and no-space scripts such as Chinese and Japanese.

\paragraph{Generation.}
Per language, we draw a deterministic random sample of up to 100{,}000 eligible sentences from the 2020--2021 data. For each item, the model receives the first half as prompt and generates a continuation. We use GPT-4.1-mini \cite{openai2025gpt41} as a representative chat-aligned model from the GPT-4 family \cite{achiam2023gpt,OpenAI_API_Reference}; at generation time, GPT-4.1 was OpenAI's frontier model. The mini variant is used to reduce inference costs for large-scale generation across 34 languages; validation checks indicate that lexical selection patterns are near-identical to those of the full model (cf.\ \S\ref{sec:validation}). Decoding is greedy (Temperature${=}0$ and Top-p${=}1$; where applicable, seeds were used). A minimal system prompt instructs the model to be a useful chat assistant. User prompts are language-specific (with informal checks by proficient speakers); see Appendix~\ref{app:prompts} for detailed prompts. A deterministic post-processing script was applied uniformly to both model continuations and human gold-standard second halves. It normalised whitespace and quotation marks, filtered digit-dominated outputs (more than 50\,\% digits) and degenerate cases such as code or HTML markup, excessive repetition, or punctuation-dominated strings.

\subsection{Metrics}
\label{sec:metrics}

\paragraph{Windowed prevalence.}
We quantify lexical usage via \emph{windowed prevalence}, as this reduces sensitivity to local repetition and improves cross-lingual comparability. For each continuation $d$, we consider a fixed-size window of $K{=}12$ tokens. With 100{,}000 prompts per language, this gives about 1.2M model tokens and 1.2M human tokens per language for estimating AI overuse. For word $w$, we define the indicator

\vspace{-0.15cm}

\begin{equation}
    I_d(w) = \indicator\bigl[w \in \text{Window}_d\bigr].
\end{equation}

\noindent The corpus-level prevalence count is then

\vspace{-0.15cm}

\begin{equation}
    c(w) = \sum_{d=1}^{N} I_d(w),
\end{equation}

\vspace{-0.15cm}

\noindent which we convert to a smoothed prevalence estimate using Jeffreys smoothing \cite{krichevsky1981performance}:

\vspace{-0.15cm}

\begin{equation}
    \ell(w) = \frac{c(w) + \tfrac{1}{2}}{N + 1}.
    \label{eq:smoothed}
\end{equation}

\vspace{-0.15cm}

\paragraph{Log Prevalence Ratio.}
We compute $\ell_H(w)$ and $\ell_M(w)$ for human and for model continuations. The \emph{Log Prevalence Ratio} (\lpr{}) is defined as

\vspace{-0.15cm}

\begin{equation}
    \text{\lpr{}}(w) = \log \frac{\ell_M(w)}{\ell_H(w)}.
    \label{eq:lpr}
\end{equation}

\noindent Positive values indicate AI overuse. To reduce instability for rare items, we require a \emph{count guard} of $c_M(w){\geq}20$ before computing \lpr{}; items below this threshold are assigned $\lpr{}{=}0$. Ranking words by \lpr{} gives per-language lists of AI-overused items. Our primary analysis focuses on \emph{content words} (NOUN, VERB, ADJ, ADV), to align our results with the literature on spiking lexical items. We also report all-word results in Appendix~\ref{app:allwordslists}. 

Jeffreys smoothing allows us to retain items with $c_H(w) = 0$ whilst avoiding undefined or infinite increases when moving from $c_H(w) = 0$ to $c_M(w) > 0$. This is particularly important because many of the strongest overrepresented items are extremely rare in the matched human text. (Note that the analyses in \S\ref{sec:prepost} and \S\ref{sec:diachronic} operate on raw WMT corpus token counts, not on smoothed values.)

We also use \emph{absolute prevalence differences}:\ $\ell_M(w) - \ell_H(w)$. \lpr{} and the absolute prevalence differences provide complementary views:\ \lpr{} highlights extreme AI-to-human ratios (focusing on spiking words, like the literature), whereas the absolute prevalence differences highlight \emph{volume shifts}.

\subsection{Pre- vs Post-GPT Analysis}
\label{sec:prepost}

For each language, we take the top 20 AI-overused content words (ranked by \lpr{}) and test whether their aggregate prevalence in news text increases from 2020--2021 to 2023--2024. As a baseline, we select 20 content words with near-zero \lpr{}, i.e.,\ words for which model and human prevalence are as similar as possible. We assess pre/post differences in aggregate prevalence using a per-language $\chi^2$ test. The $2\times2$ contingency tables contrast, within each language, counts of the top-20 AI lemma+UPOS keys against all remaining tokens in the corresponding WMT corpus period. Counts are based on raw token frequencies. As a robustness check, we apply a Bonferroni correction at $\alpha/34 \approx 0.00147$. All 34 languages remain significant under this stricter threshold, although we note that the large sample sizes likely contribute to this result. Continuations are generated with GPT-4.1-mini, for reasons of scale and cost. Because GPT-4.1-mini postdates the 2023--2024 period under study, \S\ref{sec:validation} evaluates cross-version robustness within the GPT lineage by comparing GPT-4.1-mini lists against those obtained from the temporally most relevant GPT models.

\subsection{Diachronic Analysis}
\label{sec:diachronic}

For 10 languages with extended WMT coverage (2012--2024), we track yearly prevalence (occurrences per million) of selected AI-associated content words to place the pre/post contrast in longer diachronic context. We focus on three semantic concepts (\emph{care/rigour}, \emph{emphasize}, \emph{importance}), each independently attested across multiple languages in the top-200 \lpr{} lists. For each word, we compute yearly percentage change relative to its pre-GPT baseline mean (2012--2021), which enables cross-lingual comparison on a common scale. 

\section{Results}
\label{sec:results}

\subsection{AI-Overuse Lists Across Languages}
\label{sec:overuselists}

Applying the \lpr{}-based pipeline to 34 languages gives ranked lists of AI-overused content words. Table~\ref{tab:english_top10} shows the top-10 English list for illustration. Further lists can be found in Appendix~\ref{app:overuselists}.

The English list aligns closely with prior reports on AI-associated lexicon:\ items such as \emph{align}, \emph{crucial}, and \emph{potential} have all been noted as markers of model-generated text \cite{matsui2024delving,juzek2025does,kobak2025delving}, and are recovered with our approach (cf.\ Appendix~\ref{app:scienglish}). At the same time, \lpr{} also surfaces items that are rare or unattested in human continuations but disproportionately favoured by the model, such as \emph{revolutionize} and \emph{revitalize}. This illustrates an advantage of prevalence-based ranking over manual curation:\ it recovers both widely recognised markers and less obvious ones easily missed in ad hoc inspection.

\begin{table}[t]
\centering
\small
\begin{tabular}{rlllrr}
\toprule
\textbf{\#} & \textbf{Lemma} & \textbf{POS} & \textbf{$c_H$} & \textbf{$c_M$} & \textbf{\lpr{}} \\
\midrule
1 & additionally & \textsc{adv} & 0 & 421 & 6.74 \\
2 & emphasize & \textsc{verb} & 16 & 5\,180 & 5.75 \\
3 & revolutionize & \textsc{verb} & 0 & 59 & 4.78 \\
4 & revitalize & \textsc{verb} & 0 & 44 & 4.49 \\
5 & captivated & \textsc{verb} & 1 & 94 & 4.14 \\
6 & conservationist & \textsc{noun} & 0 & 31 & 4.14 \\
7 & streamlin[e] & \textsc{verb} & 0 & 27 & 4.01 \\
8 & firsthand & \textsc{adv} & 0 & 25 & 3.93 \\
9 & personalize & \textsc{verb} & 1 & 74 & 3.91 \\
10 & introspective & \textsc{adj} & 0 & 24 & 3.89 \\
\bottomrule
\end{tabular}
\caption{Top 10 English AI-overused content words by \lpr{}. $c_H$ and $c_M$ denote human and model counts, respectively, measured in 12-token windows over 100{,}000 continuations. \lpr{} is computed with Jeffreys smoothing, so zero counts give finite values.}
\label{tab:english_top10}
\end{table}

\subsection{Cross-Lingual Semantic Convergence}
\label{sec:crosslingual}

An initial qualitative impression from the ranked lists was that overused items appeared similar across multiple languages. This suggested the possibility that AI-associated lexical overuse may show semantic convergence across languages. We assess this possibility in two ways:\ first through multilingual embedding analysis, and second through a qualitative analysis of the three highest-ranked semantic concepts across languages. 

\paragraph{Embedding-based analysis.}
Using multilingual sentence embeddings \cite{reimers2020making}, we test whether English AI-overuse seeds have unusually close semantic neighbours among the top-\(N\) \lpr{}-ranked items of other languages. For each of the top-20 English AI seeds, we compute cosine similarity to the top-\(N\) items in each target language and retain the best match. At \(N{=}200\), the AI seeds achieve a mean max-cosine of 0.730, compared to 0.668 under a UPOS-matched permutation null drawn from English content words (\(B{=}1{,}000\); \(z{=}3.29\), \(p{=}0.006\)). A matched baseline of 20 frequent, non-shifting words (smallest volume shifts, highest \(c_M\)) reaches 0.657, i.e.\ a value comparable to the null. The effect is stable across thresholds (\(N \in \{50,200,500\}\); all \(p < 0.02\)). The results are visualised in Figure~\ref{fig:cosinehook}. 

\paragraph{Qualitative analysis.}
Embedding similarity can miss genuine translation equivalents. For example, English \emph{emphasize} and Dutch \emph{beklemtonen}, despite being direct translations, only have a cosine similarity of 0.47, whilst other Dutch verbs get higher values, e.g.\ \emph{afwerken} (0.65; `to finish') or \emph{geloven} (0.50; `to believe'). Thus, we also conduct a qualitative analysis to corroborate the quantitative findings. For each language, we inspect the top-200 content words by \lpr{}, translate them into English (using Google Translate, with informal checks by proficient speakers), and assess whether prominent English AI-overuse concepts are represented among the highly ranked items. 

Table~\ref{tab:crosslingual} presents three commonly AI-overused concepts whose translation equivalents independently emerge among the top-200 AI-overused content words in many languages. Verbs expressing \emph{emphasizing}, \emph{stressing}, or \emph{highlighting} appear in 24 of 34 languages. Nouns expressing \emph{importance}, \emph{significance}, or \emph{priority} appear in 20 languages. Adjectives expressing \emph{innovative}, \emph{groundbreaking}, or \emph{cutting-edge} appear in 18 languages. 

Taken together, the embedding analysis and the qualitative examples corroborate the notion that semantically corresponding items recur across languages, including unrelated language families.

\begin{table*}[t]
\centering
\small
\begin{tabular}{lcp{11.5cm}}
\toprule
\textbf{Concept} & \textbf{$n$} & \textbf{Languages and lemmas} \\
\midrule
\makecell[l]{\textsc{emphasize / stress /}\\ \textsc{highlight} (VERB)} & 24/34 &
    AF:\ \emph{beklemtoon},\;
    AR:\ \ar{أَشَّر},\;
    BG:\ \cyr{акцентирам, подчертавам, подчертая},\;
    CS:\ \emph{zdůraznit, zdůrazňovat, podtrhovat, vyzdvihnout},\;
    DE:\ \emph{betonen, hervorheben},\;
    EL:\ \grk{υπογραμμίζω},\;
    EN:\ \emph{emphasize, highlight},\;
    ES:\ \emph{enfatizar, destacar, subrayar, realzar},\;
    ET:\ \emph{rõhutama},\;
    FI:\ \emph{korostaa, korostua},\;
    FR:\ \emph{insister, marquer},\;
    HR:\ \emph{naglasiti},\;
    ID:\ \emph{tekan, sorot},\;
    IT:\ \emph{sottolineare, evidenziare},\;
    LT:\ \emph{pabrėžti},\;
    LV:\ \emph{uzsvērt},\;
    NL:\ \emph{benadrukken},\;
    PL:\ \emph{podkreślać, podkreślić},\;
    PT:\ \emph{enfatizar, destacar, ressaltar},\;
    RO:\ \emph{sublinia, evidenția},\;
    RU:\ \cyr{подчеркивать},\;
    TR:\ \emph{vurgula},\;
    UK:\ \cyr{підкреслювати, підкреслити, наголошувати},\;
    ZH:\ \zh{强调} \\
\addlinespace
\makecell[l]{\textsc{importance / significance /}\\ \textsc{priority} (NOUN)} & 20/34 &
    AR:\ \ar{أَهَمِّيَّة},\;
    BG:\ \cyr{важност},\;
    CS:\ \emph{důležitost, důraz},\;
    DE:\ \emph{bedeutung, notwendigkeit, dringlichkeit},\;
    EL:\ \grk{σημασία},\;
    EN:\ \emph{importance},\;
    ES:\ \emph{importancia},\;
    ET:\ \emph{olulisus, võtme\_tähtsus},\;
    FA:\ \ar{اولویتبندی},\;
    FR:\ \emph{importance, nécessité},\;
    HI:\ \hi{प्राथमिकता},\;
    HR:\ \emph{važnost},\;
    ID:\ \emph{penting(nya)},\;
    IT:\ \emph{importanza},\;
    PL:\ \emph{priorytet, znaczenie},\;
    PT:\ \emph{importância},\;
    RO:\ \emph{importanță},\;
    RU:\ \cyr{важность},\;
    SR:\ \emph{važnost},\;
    UK:\ \cyr{важливість} \\
\addlinespace
\makecell[l]{\textsc{innovative /}\\ \textsc{groundbreaking} (ADJ)} & 18/34 &
    CS:\ \emph{inovativní, inovace, inovací, moderní},\;
    DE:\ \emph{innovativ, modern},\;
    EN:\ \emph{innovative, groundbreaking, advanced},\;
    ES:\ \emph{innovador},\;
    ET:\ \emph{innovaatiline},\;
    FI:\ \emph{innovatiivinen},\;
    FR:\ \emph{innovant},\;
    ID:\ \emph{inovatif},\;
    IT:\ \emph{innovativo},\;
    KK:\ \cyr{инновациялық},\;
    LT:\ \emph{inovatyvus, modernus},\;
    NL:\ \emph{innovatief},\;
    PL:\ \emph{innowacyjny, nowoczesny, zaawansowany},\;
    PT:\ \emph{inovador},\;
    RO:\ \emph{inovator, revoluționar},\;
    RU:\ \cyr{инновационный},\;
    UK:\ \cyr{інноваційний},\;
    ZH:\ \zh{先进} \\
\bottomrule
\end{tabular}
\caption{Cross-lingual semantic alignment of AI-overused content words, verified manually on the top-200 \lpr{}-ranked items per language. The $n$ column shows the number of languages (out of 34) containing at least one translation equivalent of the concept. All lemmas are drawn from independently computed, monolingual \lpr{} lists.}
\label{tab:crosslingual}
\end{table*}

\subsection{Diachronic Shift from the Pre- to the Post-GPT Period}
\label{sec:diachronicshift}

\begin{figure}[t]
    \centering
    \includegraphics[width=\columnwidth]{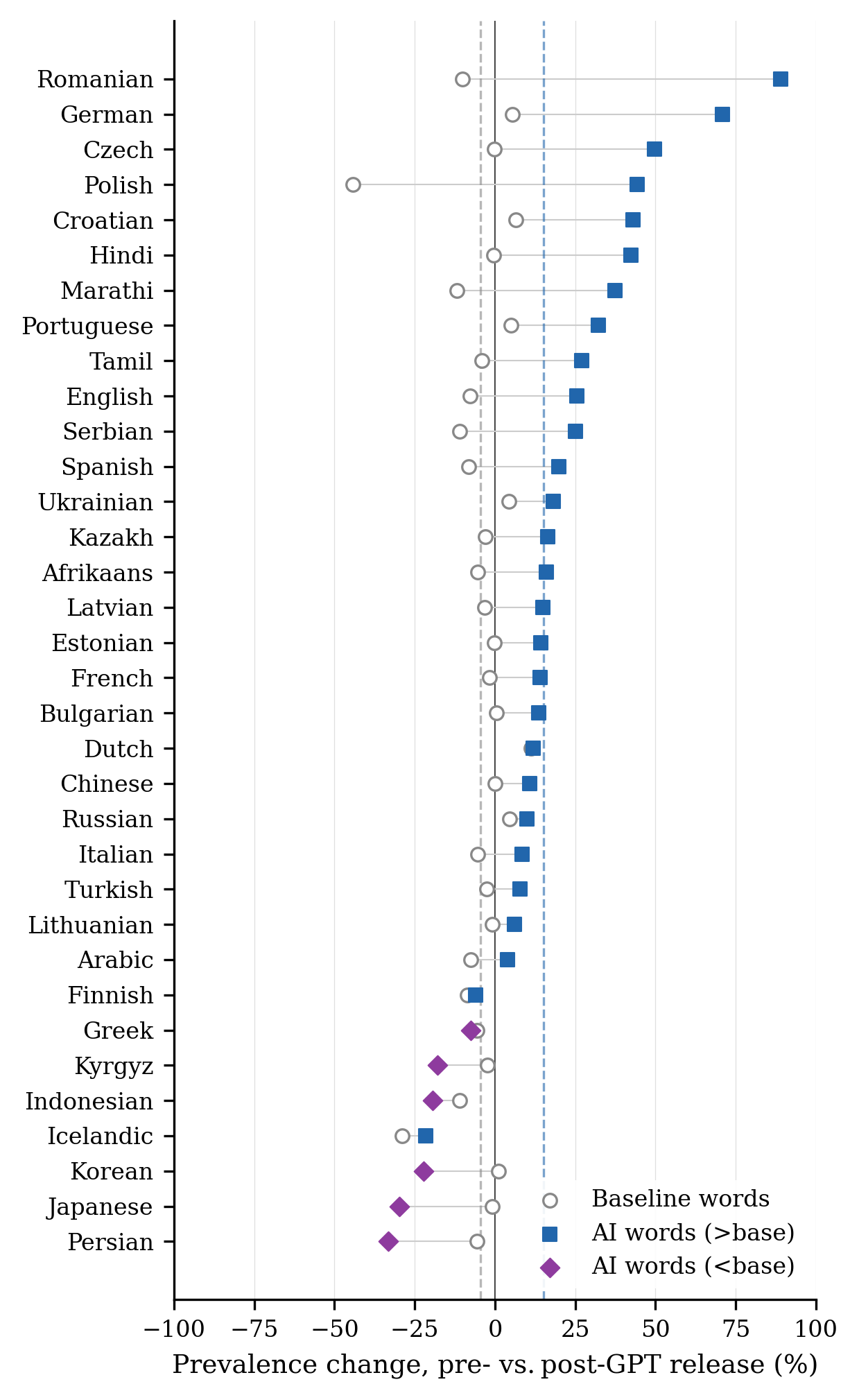}
    \caption{Prevalence change of top-20 AI-associated content words (filled squares and diamonds) vs matched baseline words (open circles) from 2020--2021 to 2023--2024, by language. Open circles denote baseline words. All AI-associated shifts are significant at $p<0.05$. Dashed lines indicate cross-language means.}
    \label{fig:marketshare}
\end{figure}

\paragraph{Content word analysis.}
Figure~\ref{fig:marketshare} shows prevalence changes from 2020--2021 to 2023--2024 for AI-associated content words (top-20 by \lpr{}) and matched baseline words across the 34 analysed languages. AI-associated words show a statistically significant shift in all 34 languages ($\chi^2$, $p < 0.05$):\ 26 languages increase and 8 decrease, with a cross-language mean change of +15.1\,\% (median:\ +14.1\,\%). By contrast, matched baseline words show a mean change of $-$4.5\,\% (median:\ $-$2.8\,\%), and in 28 of 34 languages the AI-associated set outpaces the baseline. 

The largest increases are observed in Romanian (+88.9\,\%), German (+70.8\,\%), and Czech (+49.7\,\%). English also shows a marked increase (+25.5\,\%), though smaller than the much larger disruptions reported for Scientific English. Six languages show decreases relative to their baselines (eight show absolute decreases), most notably Persian ($-$33.2\,\%), Japanese ($-$29.9\,\%), and Korean ($-$22.2\,\%). These cases may reflect variation from corpus-composition effects; we return to this in \S\ref{sec:discussion}.

\vspace{0.1cm}

\noindent\textbf{All-word analysis.}\quad Extending the analysis to all POS categories gives the same pattern, though attenuated:\ the mean change is +10.4\,\% (median:\ +11.3\,\%), compared to the matched baseline's +0.6\,\%. 

\vspace{0.1cm}

\begin{figure*}[t]
    \centering
    \includegraphics[width=\textwidth]{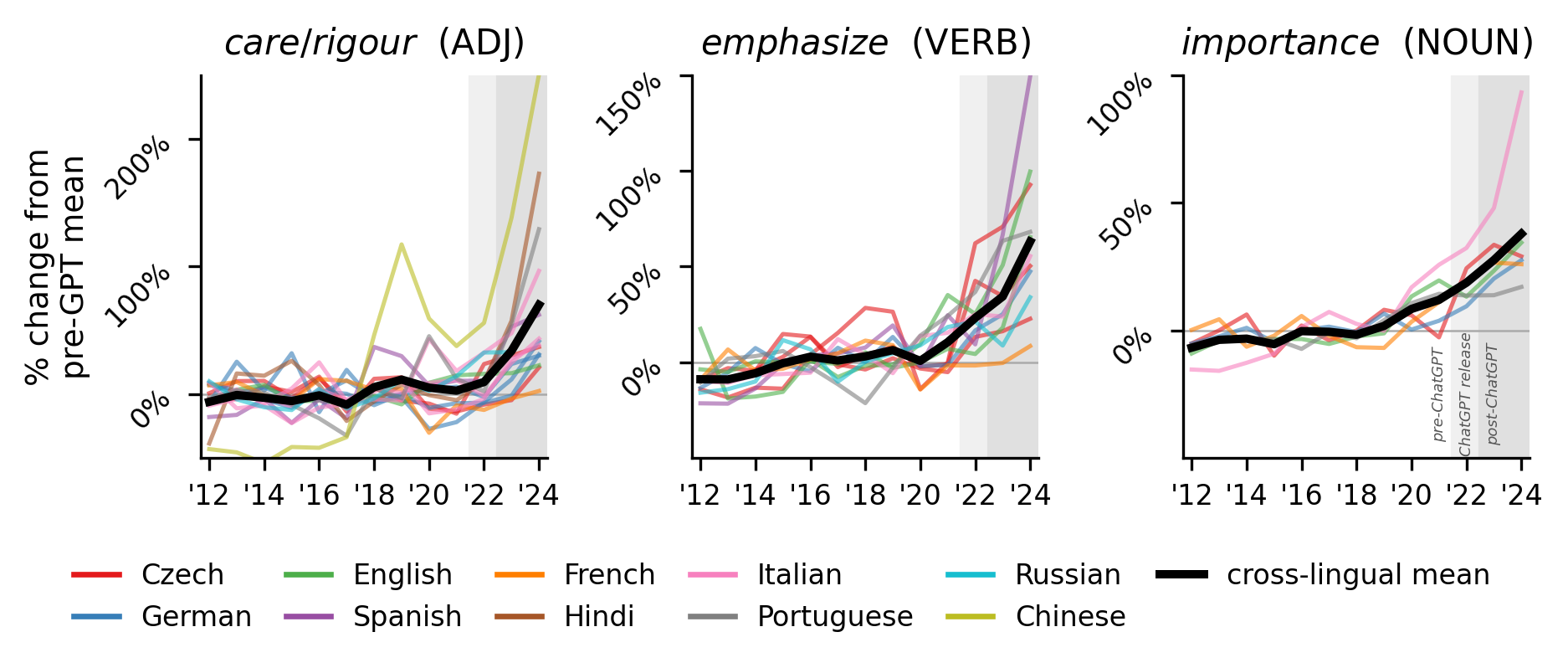}
    \caption{Longitudinal prevalence of three AI-associated semantic concepts across languages (2012--2024). Each thin line represents one language--lemma pair; the bold black line is the cross-lingual mean. Values are expressed as percentage change from the pre-GPT mean (2012--2021). Background shading distinguishes the pre-ChatGPT period (white), the year of ChatGPT's release (light grey, 2022), and the post-ChatGPT period (dark grey, 2023--2024).}
    \label{fig:diachronic}
\end{figure*}

\noindent\textbf{Longitudinal analysis.}\quad For the subset of 10 languages with WMT data going back to 2012, we track yearly prevalence (occurrences per million) for selected AI-associated content words, expressed as percentage change from their pre-GPT mean (2012--2021), over a total of about 7.1B tokens. Figure~\ref{fig:diachronic} shows three semantic concepts (\emph{care/rigour} adjectives, \emph{emphasize} verbs, and \emph{importance} nouns), each independently attested in multiple languages. Across all three panels, the cross-lingual mean remains relatively stable throughout the pre-GPT period (with slow rises possible, cf.\ \citealp{matsui2024delving}), then rises considerably after 2022. 

\section{Validation}
\label{sec:validation}

We assess robustness by rerunning the \lpr{} pipeline under alternative settings. When rankings are compared, we use Spearman correlation (\(\rho\)). Correlations are computed only over items satisfying the count guard (\(c_M{\geq}20\)) (to reduce instability induced by rare items). The reported \(n\) values indicate the number of items remaining after this filtering. Results are for English, unless otherwise stated.

\noindent\textbf{Seed robustness.}\quad Three reruns with different random seeds give highly similar lists (\(\rho{=}0.958, 0.959, 0.958\); \(n{\approx}3{,}890\) per comparison), which we treat as a baseline for natural variation. Repeating the same GPT-4.1-mini API call on the same 100k prompts under greedy decoding gives \(\rho{=}0.995\) (\(n{=}4{,}002\)), due to negligible API-level variation.

\vspace{0.1cm}

\noindent\textbf{System-prompt sensitivity.}\quad We varied the system prompt as follows, using 20{,}000 GPT-4.1-mini continuations per run at \(T{=}0\):\ (1) empty system prompt; (2) minimal task-only prompt (``Continue the input text. Output only the continuation.''); (3a) the existing system prompt vs (3b) the existing system prompt with a different seed, as a natural-variation floor. We ran this for English (Germanic), Czech (Slavic), and Spanish (Romance). Variation floor:\ \(\rho \approx 0.988\)--\(0.989\) (cs, en, es). Empty vs full:\ \(\rho{=}0.83\) (cs) and \(\rho{=}0.90\) (en, es). Minimal-task-only vs full:\ \(\rho{=}0.93\) (en) and \(\rho{=}0.95\) (cs, es). By contrast, the choice of model family matters more, see cross-architecture validation below.

\vspace{0.1cm}

\noindent\textbf{Sampling temperature.}\quad We also vary the temperature parameter with \(T \in \{0.3, 0.5, 0.7\}\) on 20{,}000 English continuations (seed 42, full system prompt), and compare these runs against two \(T{=}0\) control reruns with different seeds. Spearman \(\rho \geq 0.976\) on the top-200 \(c_M\) items across all pairwise contrasts.

\vspace{0.1cm}

\noindent\textbf{Within-GPT.}\quad Agreement across GPT-4.1 variants is high:\ mini vs nano gives \(\rho{=}0.958\) (\(n{=}3{,}708\)), which closely matches the seed baseline, and mini vs full gives \(\rho{=}0.913\) (\(n{=}3{,}762\)).

\vspace{0.1cm}

\noindent\textbf{Data-size.}\quad Relative to the default 100k-sentence run, lists from 20k, 50k, 200k, and 500k sentences remain highly similar:\ 20k (\(\rho{=}0.969\)), 50k (\(0.981\)), 200k (\(0.984\)), and 500k (\(0.977\)). The 100k default therefore lies well within a stable range.

\vspace{0.1cm}

\noindent\textbf{Window-size.}\quad Varying the prevalence window (on the same generations, post-processing) from \(K{=}10\) to \(35\) has minimal effect:\ \(K10\) (\(\rho{=}0.996\)), \(K15\) (\(0.995\)), \(K20\) (\(0.990\)), \(K25\) (\(0.986\)), \(K30\) (\(0.984\)), and \(K35\) (\(0.981\)).

\vspace{0.1cm}

\noindent\textbf{Count guard.}\quad The count guard was introduced to reduce noise. In addition to requiring \(c_M \geq 20\), we also apply a symmetric guard \(c_H \geq g\), where \(g \in \{1, 5, 10\}\). We apply this across all 34 languages and re-run the diachronic \(\chi^2\) test. Mean AI growth is \(+13.8\,\%\), \(+10.2\,\%\), and \(+11.5\,\%\), respectively (uncorrected:\ 32 of 34 languages remain significant for each $g$; Bonferroni-corrected:\ 31, 30, and 31 remain significant).

\vspace{0.1cm}

\noindent\textbf{Cross-architecture validation.}\quad Cross-model agreement is lower than within-GPT agreement:\ GPT vs Haiku gives \(\rho{=}0.85\) (\(n{=}3{,}618\)), whereas GPT vs Gemini gives \(\rho{=}0.46\) (\(n{=}3{,}017\)). This suggests a combination of shared cross-model signal and model-specific lexical preferences.

\vspace{0.1cm}

\noindent\textbf{Model-versions.}\quad Because the diachronic analysis targets 2023--2024, we compared GPT-4.1-mini lists against earlier GPT models. As some earlier models are substantially more expensive per token, the comparison was limited to 20k continuations per model. After applying the low-count guard, overlap was \(n{\approx}1{,}100\)--\(1{,}300\). Correlations are high:\ GPT-3.5 Turbo \(\rho{=}0.918\), GPT-4 \(\rho{=}0.857\), GPT-4 Turbo \(\rho{=}0.930\), and GPT-4o \(\rho{=}0.956\), the latter closely matching the seed baseline. This supports GPT-4.1-mini as a reasonable proxy for models used during the target period.

\vspace{0.1cm}

\noindent\textbf{POS-distribution-matched baseline.}\quad The headline baseline selected items with $|\lpr{}|\approx 0$. To rule out POS distribution as a driver of the AI-vs-baseline contrast, we re-ran the diachronic analysis with a baseline matched on UPOS. For each AI top-20 entry we drew one baseline lemma of the same UPOS from the $|\lpr{}|\approx 0$ pool. Across all 34 languages, mean AI growth is $+15.1\,\%$ (unchanged from the headline) against a POS-matched baseline of $-2.5\,\%$ (median $-2.9\,\%$); AI growth exceeds the matched baseline in 26 of 34 languages, and all 34 contrasts remain significant (uncorrected and Bonferroni-corrected).

\section{Discussion}
\label{sec:discussion}

Our results suggest that AI-associated lexical overuse is not confined to English, but, in parts, reflects a broader cross-lingual ``AI register''. Semantically related concepts such as \emph{emphasizing}, \emph{importance}, and \emph{innovation} recur across typologically diverse languages, consistent with accounts of reduced diversity in aligned models \cite{kirk2023understanding} and with ``shining through'' effects \cite{teich2003cross}. Plausible factors that could contribute to this convergence include:\ multilingual chat models may share a common post-training stage, they are typically built on pretraining data in which English is disproportionately prominent, and they represent different languages in a shared semantic space. Together, these properties may allow preferences learned most strongly in English or from English-dominant feedback to propagate to semantically nearby lexical choices in other languages. 

Diachronic change is clearly smaller in news than in Scientific English. One possible explanation is language proficiency:\ \citet{kobak2025delving} report stronger AI-marker uptake for authors affiliated with universities in non-English-speaking countries; at the same time, news writing is produced in contexts where writers are typically highly proficient in the language of publication. Nonetheless, the shifts observed over just a few years remain remarkable in the context of natural language change, both in their breadth and their pace. 

The longitudinal analysis indicates that many AI-associated words were already (slightly) rising before ChatGPT, and increase more sharply thereafter \cite{matsui2024delving}. This is consistent with the hypothesis that AI is amplifying pre-existing tendencies (instead of `inventing' them). It is also compatible with evidence that speakers align lexically and syntactically with artificial interlocutors \cite{brennan1991conversation,branigan2003syntactic,brandstetter2017persistent,ostrand2023rapid}. That said, the diachronic findings remain correlational, and we therefore avoid making strong causal claims.

Two qualifications are worth highlighting. First, the model-comparison analyses suggest that, although GPT-family models show strong agreement, the observed cross-lingual pattern is not uniform across model families. Haiku shares much of the GPT profile, whereas Gemini is more distinct. This strengthens the notion of model-specific lexical idiolects \cite{rudnicka2023can,rudnicka2025each}. We focus on GPT because it was the dominant chat assistant during the target period \cite{carr2024chatgptdailytraffic}. Second, the literature suggests that AI also exhibits distinct syntactic style preferences \cite{zamaraeva2025comparing}. Because our main analyses rely on log prevalence ratios, the resulting inventories are dominated by sharply overrepresented content words (as function words already tend to have high baseline prevalence in human text).

More broadly, these findings strengthen concerns that LLMs may exert homogenising pressure on human language use \cite{doshi2024generative,padmakumar2023does,moon2025homogenizing,anderson2024homogenization,kumar2025human,wenger2025we}. Our contribution to this line of work is to show that this pressure \emph{may} operate \emph{cross-linguistically}:\ the same semantic entries are promoted across dozens of languages at once (this does, however, not imply that languages are already becoming more similar overall). A plausible mechanism behind such uptake is repeated exposure:\ as users encounter AI-preferred wording both whilst interacting with assistants and whilst reading AI-assisted text, these lexical choices become increasingly familiar, easier to process, and therefore more acceptable. This possibility is in line with classic work on mere exposure \citep{zajonc1968attitudinal} and later syntheses and fluency-based accounts \citep{bornstein1989exposure,reber2004processing}. 

Several caveats remain. Six languages show decreases relative to their baseline; these cases may reflect differences in AI uptake, corpus composition (heterogeneous web-news text), or language-specific overuse profiles. Some variation is expected, given that the WMT News Crawl dataset is not of fixed composition and lacks the metadata required to control for such differences. In this respect, the results may be interpreted as a distribution centred around an overall increase in AI-overused words, with the lower tail extending into decreases. The smallest subcorpora (e.g.\ Kyrgyz, Indonesian) are the most exposed to such variation; the remaining declines may reflect language-specific circumstances, and warrant dedicated follow-ups. These limitations argue for caution, but do not undermine the central result:\ AI-associated lexical preferences form a coherent cross-lingual pattern, and in most languages that pattern is reflected in post-2022 uptake.

\section{Conclusion}
\label{sec:conclusion}

This paper examined AI-associated lexical overuse in news writing across 34 languages. We refined and extensively validated a reusable diagnostic pipeline, and have made it publicly available for future research. The pipeline allows AI-associated lexical developments to be tracked over time and may support future model development and efforts to preserve linguistic diversity.

We showed that chat-aligned models exhibit a cross-linguistically coherent lexical fingerprint:\ semantically related concepts such as \emph{emphasize}, \emph{importance}, and \emph{innovative} recur across typologically diverse languages. We further showed that these words increase in prevalence in most languages after the release of ChatGPT, whereas matched baseline words do not. Although the findings are correlational, it is plausible to conjecture that AI plays a role, given that the observed diachronic shifts are historically remarkable in both breadth and pace, and specifically concern AI-overused words.

If AI-assisted writing continues to spread, these shifts may exert unprecedented homogenising pressure on global language use. Future work could focus on causal and longitudinal follow-up analyses, particularly with respect to lower-resource languages and underrepresented registers.

\section*{Limitations}

\paragraph{Historical model matching.}
The main lexical lists are derived from GPT-4.1-mini, whereas the uptake window examined in the corpus is 2023--2024, when earlier GPT models such as GPT-3.5 Turbo, GPT-4, GPT-4 Turbo, and GPT-4o were current. This choice is motivated by computational cost (see Appendix~\ref{app:compute}). Under the hypothesis that the observed longitudinal shifts are partly influenced by AI, GPT-4.1-mini is therefore not treated as the direct source of uptake, but rather as a scalable proxy for the dominant models of the period.

\paragraph{Training-data overlap.}
The pre-GPT reference period (2020--2021) plausibly overlaps with the training data of the models used here. This complicates interpretation, since such overlap could either dampen apparent change by making model preferences closer to the pre-period baseline, or strengthen correspondence if the model amplifies stylistic tendencies already present in its training distribution. Without detailed knowledge of model training corpora, the direction and magnitude of this effect cannot be determined precisely.

\paragraph{Single-model primary generation.}
The primary generation pipeline uses a single model, GPT-4.1-mini, across all 34 languages, with cross-architecture checks on Claude Haiku and Gemini Flash (\S\ref{sec:validation}). The resulting lists are therefore necessarily GPT-centric; this choice is motivated by computational costs (see Appendix~\ref{app:compute}). GPT-family models are ecologically well motivated for the 2023--2024 period, as they dominated the market \cite{carr2024chatgptdailytraffic}. However, stronger claims about cross-model universality would require full multi-model lists for each language.

\paragraph{Corpus representativeness.}
Our diachronic analysis is based on WMT News Crawl, which captures heterogeneous web-news. The data are given in the form of shuffled and deduplicated sentences. Source composition varies across languages and years, so some observed shifts may reflect changes in the underlying source mix (instead of population-level language change). Critically, we are unable to control for this due to the lack of per-sentence metadata, which is one motivation for the use of baseline items. The results should therefore be interpreted as characterising the web-news register represented in WMT News Crawl, not news writing in a fully general sense.

\paragraph{Languages with decreases.}
Six languages show decreases in AI-associated lexical prevalence after 2022 relative to their baselines, eight languages show absolute decreases. These cases may reflect differences in AI uptake, variation due to shifts in corpus composition over time (see the above note on corpus composition), and/or language-specific properties of the measured overuse profile. 

\paragraph{Temporal coverage.}
Our diachronic analysis ends in 2024. We could not include 2025 because comparable 2025 WMT News Crawl data were not yet available in a stable form across languages at the time of analysis. 

\paragraph{POS tagging and lemmatisation artefacts.}
All analyses depend on automatic tokenisation, POS tagging, and lemmatisation with Stanza, which introduces occasional inconsistencies. Examples include German noun capitalisation variants, English truncation of final \emph{-e}, and alternative Russian lemma forms involving \cyr{ё}/\cyr{е}. We apply normalisation where possible, but residual mismatches may cause a small number of items to be undercounted. No-space scripts (Chinese, Japanese) and morphologically rich languages (e.g.\ Korean, Persian) may introduce additional uncertainty. We rely on Stanza's performance; ideally, tagging quality would be spot-checked across script types.

\paragraph{Semantic convergence:\ alternative approaches.}
Our approaches to cross-lingual convergence rely on multilingual embedding analysis and qualitative analysis of the highest-ranked semantic concepts. However, multilingual embeddings can be relatively crude, as illustrated by the earlier \emph{emphasize}/\emph{beklemtonen} example, where two near-direct semantic equivalents did not emerge as strongly similar. Likewise, the qualitative analysis may be sensitive to variation across replications. An alternative avenue for a more quantitative and reproducible analysis could involve the use of BabelNet \cite{navigli2012babelnet,bevilacqua2020breaking}. We note, however, that BabelNet coverage may be limited for some of the lower-resource languages in our sample.

\paragraph{Lexical scope.}
The present study focuses on lexical prevalence. This choice reflects a trade-off between scope and comparability:\ lexical frequency is the most robust signal available across 34 languages in heterogeneous news corpora. Broader linguistic measures of homogenisation, e.g.\ syntactic, semantic, or discourse-level convergence, remain an important target for future work.

\paragraph{Spikes vs volume shifts, top-50.}
The diachronic shifts reported here concern a relatively small inventory of `spiking' words (per language, the top-20 AI-overused content words). If all parts of speech are included in the top-20, the mean increase is $+10.4\,\%$ (down from $+15.1\,\%$). Furthermore, rather than measuring shifts in spiking words via $|\lpr{}|$, an alternative approach would be to measure shifts based on raw frequency alone (i.e.,\ volume shifts). We are therefore cautious about making broader claims regarding changes in language use. If the Top-\(N\) cutoff is changed to top-50 (same protocol, content-only), the cross-language mean AI shift becomes \(+9.9\,\%\) (vs \(+15.1\,\%\) at top-20), with a mean baseline shift of \(-3.5\,\%\). Furthermore, the top-50 overused words exceed a top-50 item baseline in 25 of 34 languages (uncorrected:\ all significant; Bonferroni-corrected:\ 32 of 34). The choice of focusing on the top-20 content words was made to connect the present analysis to the existing literature on AI-overused words.

\section*{Ethical Considerations}

\paragraph{Potential risks.}
Our results are primarily descriptive, and the findings should not be overgeneralised into normative claims about language use or applied in authorship-screening settings. This is particularly relevant in the context of AI detection. Existing AI detectors have been shown to exhibit negative bias against non-native speakers \cite{liang2023gpt}, which can cause harm in high-stakes contexts such as academic integrity investigations or hiring decisions. Our work is not intended for AI detection, but rather for the identification of broader AI language behaviour. Lexical signals for AI detection can be unstable \cite{schmalz2025can}, which is consistent with broader findings regarding reliability issues of AI detection systems \cite{weber2023testing}. Using lexical signals for AI detection carries the risk of stigmatising non-standard language varieties, especially those associated with non-native speakers. 

If the outlined homogenisation effects are real, it is plausible that they will be strongest for lower-resource languages and underrepresented registers due to their more limited computational representation. One possible consequence could be a loss of linguistic diversity. This underlines the need for future work on the impact of AI on lower-resource languages and underrepresented registers, as well as for, more broadly, the inclusion of these languages and registers in the development of computational tools.

\paragraph{Data sensitivity.}
We use a pre-existing large-scale news corpus and did not conduct a dedicated manual check for personally identifying or offensive content beyond the dataset's existing curation. Further, our analysis is performed at the aggregate lexical level.

\paragraph{AI usage.}
The author wrote the paper. AI assistants (GPT, Gemini, and Claude) were used to remove language errors and for improving wording. Code development was AI-assisted (same models); all code was reviewed and tested. 

\section*{Acknowledgements}

We thank Gordon Erlebacher for valuable input throughout the project, and Zina Ward for discussions on the broader research line. Computational support for the precursor project was kindly provided by Jose Hernandez and the FSU Research Computing Center. We are also grateful to the reviewers for their constructive feedback.

\bibliography{references}

\appendix

\section*{Appendices}

\section{Code, Data, Computational Setup}
\label{app:compute}

\paragraph{Code and Data.} 
All code, with notes on how to retrieve data, is available at:\ \href{https://github.com/tjuzek/ai-34-languages}{github.com/tjuzek/ai-34-languages}. The repository includes an interactive web-based explorer for all 34 language lists, together with instructions for local hosting. The explorer can also be accessed at \href{https://www.aiwordexplorer.com/}{aiwordexplorer.com/}. It supports browsing by language, POS category, and model, and displays \lpr{}, volume shifts, and per-million-word prevalence for each item.

\paragraph{Computational Set-up.} All major computations were run on a machine with the following specifications:

\vspace{0.1cm}

\noindent\textbf{(A) GPU server.}
NVIDIA H100 PCIe (80\,GB); driver~570.148.08; CUDA~12.8. 
Intel Xeon Platinum 8480+; 221\,GiB RAM. 
Ubuntu~24.04.2 LTS; Linux~6.11.0-29.

\vspace{0.1cm}

\noindent\textbf{Software.}
Python~3.12.3; PyTorch~2.8.0{+}cu128 (CUDA~12.8; cuDNN~91002); 
\texttt{transformers}~4.56.1; \texttt{accelerate}~1.10.1; 
\texttt{peft}~0.17.1; \texttt{stanza}~1.11.0; \texttt{sentence-transformers} using \texttt{paraphrase-multilingual-MiniLM-L12-v2}. 

\vspace{0.1cm}

\noindent\textbf{Costs.} Total computation time was around 460 hours, the majority of which was spent on part-of-speech tagging. API generation costs for GPT-4.1-mini (approximately 100k continuations per language across 34 languages) totalled approximately \$1,100.

\section{Generation Prompts}
\label{app:prompts}

\paragraph{System prompt.}
\begin{quote}
\small\ttfamily
You are a helpful and knowledgeable assistant.
Follow the user's instructions and focus on the task they provide.
Your task is to provide the immediate continuation of the provided text fragment.

Guidelines: \\
1. Do NOT repeat the input text. \\
2. If the input text is machine-formatted or technical data, output an empty string ("") and nothing else. \\
3. Do NOT provide any conversational preface or acknowledgments (e.g.,\ "Here is the continuation..."). \\
4. Output ONLY natural language. \\
5. Output ONLY the continuation text.
\end{quote}

\paragraph{User prompts.}
\begin{quote}
\small\ttfamily
English: Provide a continuation of this English news text, without preamble, continue directly:\textbackslash n\textbackslash n[\textsc{prompt text}]
\end{quote}

\begin{quote}
\small\ttfamily
Dutch: Schrijf het vervolg van deze Nederlandse nieuwstekst, zonder inleiding, ga direct door:\textbackslash n\textbackslash n[\textsc{prompt text}]
\end{quote}

\noindent Language-specific user prompts are provided in the GitHub repository.

\section{Diachronic Word Lists}
\label{app:diachronic}

Table~\ref{tab:diachronic_words} lists the language--lemma pairs used in the longitudinal analysis shown in Figure~\ref{fig:diachronic}. Words were selected from the top-200 content words by \lpr{} in each language and grouped by semantic concept.

\begin{table}[h]
\centering
\small
\begin{tabular}{lll}
\toprule
\textbf{Lang} & \textbf{Lemma} & \textbf{Translation} \\
\midrule
\multicolumn{3}{l}{\textsc{importance} (NOUN)} \\
\cmidrule(lr){1-3}
cs & \emph{důležitost} & importance \\
de & \emph{Bedeutung} & significance \\
en & \emph{importance} & importance \\
fr & \emph{importance} & importance \\
it & \emph{importanza} & importance \\
pt & \emph{importância} & importance \\
\midrule
\multicolumn{3}{l}{\textsc{emphasize} (VERB)} \\
\cmidrule(lr){1-3}
cs & \emph{zdůraznit} & emphasize \\
cs & \emph{zdůrazňovat} & stress \\
cs & \emph{vyzdvihnout} & highlight \\
de & \emph{hervorheben} & emphasize \\
en & \emph{emphasize} & emphasize \\
en & \emph{highlight} & highlight \\
es & \emph{realzar} & highlight \\
fr & \emph{marquer} & mark \\
it & \emph{evidenziare} & highlight \\
pt & \emph{enfatizar} & emphasize \\
ru & \cyr{подчеркивать} & emphasize \\
\midrule
\multicolumn{3}{l}{\textsc{care / rigour} (ADJ)} \\
\cmidrule(lr){1-3}
cs & \emph{pečlivý} & careful \\
cs & \emph{precizní} & precise \\
de & \emph{präzis[en]} & precise \\
de & \emph{sorgfältig} & careful \\
en & \emph{thorough} & thorough \\
es & \emph{impecable} & impeccable \\
fr & \emph{rigide} & rigid \\
hi & \hi{समग्र} & comprehensive \\
it & \emph{mirato} & targeted \\
it & \emph{impeccabile} & impeccable \\
it & \emph{approfondito} & in-depth \\
pt & \emph{rigoroso} & rigorous \\
ru & \cyr{внимательный} & attentive \\
zh & \zh{精湛} & exquisite \\
\bottomrule
\end{tabular}
\caption{Word lists used in the longitudinal analysis shown in Figure~\ref{fig:diachronic}, grouped by semantic concept.}
\label{tab:diachronic_words}
\end{table}

\section{Representative Absolute Diachronic Shifts}
\label{app:opm}

The relative shifts reported in \S\ref{sec:diachronicshift} correspond to substantial shifts in absolute frequency. A sensible way to normalise these frequencies is to express them as occurrences per million tokens (OPM). Table~\ref{tab:opm} lists representative English AI-overused items from the diachronic comparison, contrasting OPM values for 2020--2021 and 2023--2024.

\begin{table}[h]
\centering
\small
\begin{tabular}{lrrr}
\toprule
\textbf{Item} & \textbf{OPM 2020/1} & \textbf{2023/4} & \textbf{\% change} \\
\midrule
\emph{additionally} (A)   & 15.21 & 22.62 & $+$48.7 \\
\emph{emphasize} (V)     & 17.94 & 25.69 & $+$43.2 \\
\emph{captivated} (V)    &  1.37 &  2.34 & $+$70.8 \\
\emph{importance} (N)    & 39.86 & 44.13 & $+$10.7 \\
\emph{resilience} (N)    & 14.69 & 18.11 & $+$23.3 \\
\emph{dedication} (N)    &  7.40 & 12.01 & $+$62.3 \\
\emph{revolutionize} (V) &  1.38 &  1.59 & $+$15.2 \\
\bottomrule
\end{tabular}
\caption{Representative English AI-overused items in the diachronic comparison:\ mean occurrences per million tokens (OPM) for 2020--2021 and 2023--2024, together with percentage change.}
\label{tab:opm}
\end{table}

\section{Scientific English Comparison}
\label{app:scienglish}

For comparison with prior work on Scientific English, we use the list of AI-overused items reported by \citet{galpin2025exploring}:
\texttt{advancement\_NOUN}, \texttt{align\_VERB}, \texttt{boast\_VERB}, \texttt{commendable\_ADJ}, \texttt{comprehend\_VERB}, \texttt{crucial\_ADJ}, \texttt{delve\_VERB}, \texttt{emphasize\_VERB}, \texttt{garner\_VERB}, \texttt{groundbreaking\_ADJ}, \texttt{intricacy\_NOUN}, \texttt{intricate\_ADJ}, \texttt{invaluable\_ADJ}, \texttt{meticulous\_ADJ}, \texttt{meticulously\_ADV}, \texttt{notable\_ADJ}, \texttt{noteworthy\_ADJ}, \texttt{pivotal\_ADJ}, \texttt{potential\_ADJ}, \texttt{potential\_NOUN}, \texttt{realm\_NOUN}, \texttt{showcase\_VERB}, \texttt{showcase\_NOUN}, \texttt{significant\_ADJ}, \texttt{strategically\_ADV}, \texttt{surpass\_VERB}, and \texttt{underscore\_VERB}.

Of these 27 entries, 22 are likewise identified as AI-overused in our English news data, indicating substantial, though not complete, overlap with previously reported Scientific English patterns and the lexical shifts observed here. The five items not attested as AI-overused in our English news data are \texttt{comprehend\_VERB}, \texttt{noteworthy\_ADJ}, \texttt{realm\_NOUN}, \texttt{showcase\_NOUN}, and \texttt{surpass\_VERB}. 

\section{Selected AI-Overuse Lists}
\label{app:overuselists}

\subsection{Selected Content-Word Lists}
\label{app:contentwords}

Tables~\ref{tab:spanish_top10} and \ref{tab:french_top10} show the top-10 AI-overused content words by \lpr{} for Spanish and French. 

\begin{table}[H]
\centering
\small
\begin{tabular}{rlllrr}
\toprule
\textbf{\#} & \textbf{Lemma} & \textbf{POS} & \textbf{$c_H$} & \textbf{$c_M$} & \textbf{\lpr{}} \\
\midrule
1 & testigos & \textsc{noun} & 0 & 146 & 5.68 \\
2 & organizaciones & \textsc{noun} & 0 & 123 & 5.51 \\
3 & imborrable & \textsc{adj} & 0 & 68 & 4.92 \\
4 & estudios & \textsc{noun} & 0 & 59 & 4.78 \\
5 & analistas & \textsc{noun} & 0 & 41 & 4.42 \\
6 & equipos & \textsc{noun} & 0 & 36 & 4.29 \\
7 & multidisciplinario & \textsc{adj} & 0 & 28 & 4.04 \\
8 & reinserción & \textsc{noun} & 0 & 27 & 4.01 \\
9 & empresas & \textsc{noun} & 0 & 26 & 3.97 \\
10 & intensificación & \textsc{noun} & 0 & 23 & 3.85 \\
\bottomrule
\end{tabular}
\caption{Top 10 Spanish AI-overused content words by \lpr{}. $c_H$ and $c_M$ denote human and model counts, respectively, measured in 12-token windows over 100{,}000 continuations.}
\label{tab:spanish_top10}
\end{table}

\begin{table}[H]
\centering
\small
\begin{tabular}{rlllrr}
\toprule
\textbf{\#} & \textbf{Lemma} & \textbf{POS} & \textbf{$c_H$} & \textbf{$c_M$} & \textbf{\lpr{}} \\
\midrule
1 & captiver & \textsc{verb} & 0 & 139 & 5.63 \\
2 & accru & \textsc{verb} & 0 & 34 & 4.23 \\
3 & géopolitique & \textsc{adj} & 4 & 275 & 4.11 \\
4 & habilement & \textsc{adv} & 1 & 84 & 4.03 \\
5 & impartialité & \textsc{noun} & 0 & 24 & 3.89 \\
6 & dualité & \textsc{noun} & 0 & 22 & 3.81 \\
7 & poignant & \textsc{adj} & 1 & 64 & 3.76 \\
8 & compromettant & \textsc{verb} & 0 & 21 & 3.76 \\
9 & réévaluation & \textsc{noun} & 0 & 21 & 3.76 \\
10 & palpable & \textsc{adj} & 3 & 129 & 3.61 \\
\bottomrule
\end{tabular}
\caption{Top 10 French AI-overused content words by \lpr{}. $c_H$ and $c_M$ denote human and model counts, respectively, measured in 12-token windows over 100{,}000 continuations.}
\label{tab:french_top10}
\end{table}

\subsection{Selected All-Word Lists}
\label{app:allwordslists}

Tables~\ref{tab:english_top10_all} and \ref{tab:english_volshift_top10} show the top-10 English items by \lpr{} and by volume shift (all POS categories).

\begin{table}[H]
\centering
\small
\begin{tabular}{rlllrr}
\toprule
\textbf{\#} & \textbf{Lemma} & \textbf{POS} & \textbf{$c_H$} & \textbf{$c_M$} & \textbf{\lpr{}} \\
\midrule
1 & additionally & \textsc{adv} & 0 & 421 & 6.74 \\
2 & emphasize & \textsc{verb} & 16 & 5\,180 & 5.75 \\
3 & revolutionize & \textsc{verb} & 0 & 59 & 4.78 \\
4 & revitalize & \textsc{verb} & 0 & 44 & 4.49 \\
5 & renforcer & \textsc{x} & 0 & 36 & 4.29 \\
6 & captivated & \textsc{verb} & 1 & 94 & 4.14 \\
7 & conservationist & \textsc{noun} & 0 & 31 & 4.14 \\
8 & streamlin[e] & \textsc{verb} & 0 & 27 & 4.01 \\
9 & firsthand & \textsc{adv} & 0 & 25 & 3.93 \\
10 & personalize & \textsc{verb} & 1 & 74 & 3.91 \\
\bottomrule
\end{tabular}
\caption{Top 10 English AI-overused items by \lpr{} (all categories). $c_H$ and $c_M$ denote human and model counts, respectively, measured in 12-token windows over 100{,}000 continuations.}
\label{tab:english_top10_all}
\end{table}

\begin{table}[H]
\centering
\small
\begin{tabular}{rlllrr}
\toprule
\textbf{\#} & \textbf{Lemma} & \textbf{POS} & \textbf{$c_H$} & \textbf{$c_M$} & \textbf{Vol.\ shift} \\
\midrule
1 & and & \textsc{cconj} & 32\,242 & 41\,456 & 9\,214 \\
2 & the & \textsc{det} & 47\,458 & 54\,692 & 7\,234 \\
3 & this & \textsc{det} & 2\,673 & 8\,419 & 5\,746 \\
4 & emphasize & \textsc{verb} & 16 & 5\,180 & 5\,164 \\
5 & that & \textsc{sconj} & 4\,149 & 8\,254 & 4\,105 \\
6 & have & \textsc{aux} & 7\,710 & 10\,892 & 3\,182 \\
7 & importance & \textsc{noun} & 68 & 2\,684 & 2\,616 \\
8 & authority & \textsc{noun} & 230 & 2\,818 & 2\,588 \\
9 & aim & \textsc{verb} & 112 & 2\,664 & 2\,552 \\
10 & health & \textsc{noun} & 797 & 3\,278 & 2\,481 \\
\bottomrule
\end{tabular}
\caption{Top 10 English items by (rescaled) volume shift ($c_M - c_H$). $c_H$ and $c_M$ denote human and model counts, respectively, measured in 12-token windows over 100{,}000 continuations. For ease of interpretation, we give a linear rescaling of the absolute prevalence difference $\ell_M(w) - \ell_H(w)$ introduced in \S\ref{sec:metrics}.}
\label{tab:english_volshift_top10}
\end{table}

\noindent Additional lists for all languages are available in our GitHub repository (cf.\ Appendix~\ref{app:compute}).

\end{document}